\documentclass[11pt]{article}

\usepackage{sonto-paper}

\usepackage[british]{babel}

\usepackage{tabularx}
\usepackage{multirow}
\usepackage{amsmath}
\usepackage{amssymb}
\usepackage{siunitx}
\usepackage{placeins}
\usepackage{tikz}
\usepackage{pgfplots}
\pgfplotsset{compat=1.18}

\usepackage{hyperref}
\usepackage{url}
\usepackage[nameinlink,noabbrev]{cleveref}

\hypersetup{
    pdftitle={Frontier Financial Judgement: Can agents tell what might move a stock?},
    pdfauthor={Joshua Harris}
}

\graphicspath{{figures/}}
\newcommand{\CompanyA}{Company A}
\newcommand{\CompanyB}{Company B}
\newcommand{\CompanyC}{Company C}
\newcommand{\CompanyD}{Company D}
\newcommand{\CompetitorA}{Competitor A}

\title{Frontier Financial Judgement: Can agents tell what might move a stock?}
\author{Sonto\smash{\ \textbar{}\ {\fontsize{12}{15}\selectfont\mdseries\color{black!45}Joshua Harris}}}
\date{22 July 2026}
\sontologo{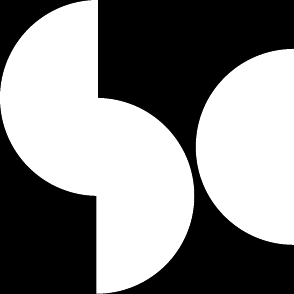}
\sontocorrespondence{\href{mailto:joshua.harris@sontoresearch.ai}{joshua.harris@sontoresearch.ai}}
\sontocorrespondencemark{\vphantom{\textsuperscript{*}}\smash{{\fontsize{12}{15}\selectfont\mdseries\color{black!45}\textsuperscript{*}}}}

\begin{document}

\maketitle

\begin{abstract}
We introduce Frontier Financial Judgement, a challenging new benchmark developed in collaboration with professional equity analysts to assess agents' ability to replicate expert human judgements. Rapidly identifying new information, evaluating its implications and determining its valuation impact is one of the most time-consuming and challenging aspects of real-world equity coverage. This is becoming ever more difficult and important as AI rapidly increases the quantity of new information to process. The strongest agent we evaluate on Frontier Financial Judgement matches all expert labels in only 52.4\% of cases. We also find significant divergence in estimated false-positive rates among frontier agents, ranging from \textasciitilde{}1\% for GPT-5.6 Sol to \textasciitilde{}32\% for Claude Sonnet 4.6. To construct the benchmark and make it representative of real-world settings, we combine human-designed and labelled synthetic articles with live news articles and historical documents, creating 656 items for assessment. The resulting task requires agents to distinguish genuinely new, valuation-relevant financial information from stale, immaterial or misleading news under realistic conditions. We find substantial trade-offs among agent accuracy, cost, false positives and reliability that continue to hinder the reliable deployment of news-flow filtering in practice.
\end{abstract}


\begin{figure}[!h]
    \centering
    \begingroup
    \definecolor{tbchartbest}{HTML}{E2F3E5}
    \definecolor{tbchartcool}{HTML}{F2F5F6}
    \definecolor{tbchartwarm}{HTML}{F7F4F0}
    \definecolor{tbchartsoftred}{HTML}{F8EEEE}
    \definecolor{tbchartgpt}{HTML}{2D6A8A}
    \definecolor{tbchartclaude}{HTML}{C96A55}
    \definecolor{tbchartdeepseek}{HTML}{3D8B68}
    \definecolor{tbchartmimo}{HTML}{DF922D}
    \definecolor{tbchartqwen}{HTML}{8467A9}
    \definecolor{tbchartnemotron}{HTML}{5F8F3A}
    \begin{tikzpicture}
        \begin{axis}[
            width=\linewidth,
            height=7.20cm,
            xmode=log,
            xmin=0.611, xmax=300.0,
            ymin=2.8, ymax=72.8,
            xtick={1,3,10,30,100,300},
            xticklabels={\$1,\$3,\$10,\$30,\$100,\$300},
            ytick={10,20,30,40,50,60,70},
            xlabel={Cost (USD, log scale)},
            ylabel={All-label accuracy (\%)},
            axis line style={draw=tbmuted!65, line width=0.45pt},
            tick style={draw=tbmuted!65},
            tick label style={font=\scriptsize, text=tbmuted},
            label style={font=\small, text=tbink},
            major grid style={draw=white, line width=0.8pt},
            grid=major,
            set layers=standard,
            clip=false,
        ]
            \addplot[on layer=axis background, draw=none, fill=tbchartbest, forget plot] coordinates {(0.611,37.805) (13.534,37.805) (13.534,72.805) (0.611,72.805)} \closedcycle;
            \addplot[on layer=axis background, draw=none, fill=tbchartcool, forget plot] coordinates {(13.534,37.805) (300.000,37.805) (300.000,72.805) (13.534,72.805)} \closedcycle;
            \addplot[on layer=axis background, draw=none, fill=tbchartwarm, forget plot] coordinates {(0.611,2.805) (13.534,2.805) (13.534,37.805) (0.611,37.805)} \closedcycle;
            \addplot[on layer=axis background, draw=none, fill=tbchartsoftred, forget plot] coordinates {(13.534,2.805) (300.000,2.805) (300.000,37.805) (13.534,37.805)} \closedcycle;
            \addplot[draw=tbmuted!65, dashed, line width=0.6pt, forget plot] coordinates {(13.534,2.805) (13.534,72.805)};
            \addplot[draw=tbmuted!65, dashed, line width=0.6pt, forget plot] coordinates {(0.611,37.805) (300.000,37.805)};
            \node[anchor=north west, font=\bfseries\scriptsize, text=tbchartdeepseek] at (rel axis cs:0.018,0.975) {MOST ATTRACTIVE};
            \node[anchor=north east, font=\bfseries\tiny, text=tbmuted] at (rel axis cs:0.982,0.975) {HIGHER ACCURACY, HIGHER COST};
            \node[anchor=south west, font=\bfseries\tiny, text=tbmuted] at (rel axis cs:0.018,0.025) {LOWER COST, LOWER ACCURACY};
            \node[anchor=south east, font=\bfseries\tiny, text=tbmuted] at (rel axis cs:0.982,0.025) {LEAST ATTRACTIVE};
            \addplot[only marks, mark=*, mark size=3.15pt, draw=white, fill=tbchartdeepseek, line width=0.9pt, forget plot] coordinates {(0.973458,34.146341)};
            \node[anchor=west, font=\scriptsize, text=tbink, xshift=5.0pt, yshift=7.0pt] at (axis cs:0.973458,34.146341) {DeepSeek V4 Flash};
            \addplot[only marks, mark=*, mark size=3.15pt, draw=white, fill=tbchartmimo, line width=0.9pt, forget plot] coordinates {(1.537489,32.926829)};
            \node[anchor=west, font=\scriptsize, text=tbink, xshift=5.0pt, yshift=-8.0pt] at (axis cs:1.537489,32.926829) {MiMo-V2.5};
            \addplot[only marks, mark=*, mark size=3.15pt, draw=white, fill=tbchartdeepseek, line width=0.9pt, forget plot] coordinates {(3.149886,40.243902)};
            \node[anchor=west, font=\scriptsize, text=tbink, xshift=5.0pt, yshift=5.0pt] at (axis cs:3.149886,40.243902) {DeepSeek V4 Pro};
            \addplot[only marks, mark=*, mark size=3.15pt, draw=white, fill=tbchartqwen, line width=0.9pt, forget plot] coordinates {(3.695502,21.951220)};
            \node[anchor=east, font=\scriptsize, text=tbink, xshift=-5.0pt, yshift=8.0pt] at (axis cs:3.695502,21.951220) {Qwen3.6 35B-A3B};
            \addplot[only marks, mark=*, mark size=3.15pt, draw=white, fill=tbchartqwen, line width=0.9pt, forget plot] coordinates {(6.516128,20.731707)};
            \node[anchor=west, font=\scriptsize, text=tbink, xshift=5.0pt, yshift=-7.0pt] at (axis cs:6.516128,20.731707) {Qwen3.7 Plus};
            \addplot[only marks, mark=*, mark size=3.15pt, draw=white, fill=tbchartnemotron, line width=0.9pt, forget plot] coordinates {(8.540247,8.536585)};
            \node[anchor=west, font=\scriptsize, text=tbink, xshift=5.0pt, yshift=8.0pt] at (axis cs:8.540247,8.536585) {Nemotron 3 Super};
            \addplot[only marks, mark=*, mark size=3.15pt, draw=white, fill=tbchartgpt, line width=0.9pt, forget plot] coordinates {(12.011930,42.682927)};
            \node[anchor=east, font=\scriptsize, text=tbink, xshift=-5.0pt, yshift=6.0pt] at (axis cs:12.011930,42.682927) {GPT-5.6 Luna};
            \addplot[only marks, mark=*, mark size=3.15pt, draw=white, fill=tbchartgpt, line width=0.9pt, forget plot] coordinates {(15.056437,30.487805)};
            \node[anchor=west, font=\scriptsize, text=tbink, xshift=5.0pt, yshift=-7.0pt] at (axis cs:15.056437,30.487805) {GPT-5.4 mini};
            \addplot[only marks, mark=*, mark size=3.15pt, draw=white, fill=tbchartgpt, line width=0.9pt, forget plot] coordinates {(25.632333,46.341463)};
            \node[anchor=west, font=\scriptsize, text=tbink, xshift=5.0pt, yshift=4.0pt] at (axis cs:25.632333,46.341463) {GPT-5.6 Terra};
            \addplot[only marks, mark=*, mark size=3.15pt, draw=white, fill=tbchartgpt, line width=0.9pt, forget plot] coordinates {(41.752417,40.243902)};
            \node[anchor=east, font=\scriptsize, text=tbink, xshift=-5.0pt, yshift=7.0pt] at (axis cs:41.752417,40.243902) {GPT-5.4};
            \addplot[only marks, mark=*, mark size=3.15pt, draw=white, fill=tbchartgpt, line width=0.9pt, forget plot] coordinates {(76.338320,52.439024)};
            \node[anchor=east, font=\scriptsize, text=tbink, xshift=-5.0pt, yshift=8.0pt] at (axis cs:76.338320,52.439024) {GPT-5.5};
            \addplot[only marks, mark=*, mark size=3.15pt, draw=white, fill=tbchartgpt, line width=0.9pt, forget plot] coordinates {(83.553491,51.219512)};
            \node[anchor=west, font=\scriptsize, text=tbink, xshift=5.0pt, yshift=-8.0pt] at (axis cs:83.553491,51.219512) {GPT-5.6 Sol};
            \addplot[only marks, mark=*, mark size=3.15pt, draw=white, fill=tbchartclaude, line width=0.9pt, forget plot] coordinates {(110.515047,36.585366)};
            \node[anchor=north, font=\scriptsize, text=tbink, xshift=0.0pt, yshift=-6.0pt] at (axis cs:110.515047,36.585366) {Claude Sonnet 4.6};
            \addplot[only marks, mark=*, mark size=3.15pt, draw=white, fill=tbchartclaude, line width=0.9pt, forget plot] coordinates {(206.456420,39.024390)};
            \node[anchor=south, font=\scriptsize, text=tbink, xshift=0.0pt, yshift=6.0pt] at (axis cs:206.456420,39.024390) {Claude Opus 4.8};
        \end{axis}
    \end{tikzpicture}
    \endgroup
    \caption{Cost and strict all-label accuracy on Frontier Financial Judgement.}\label{fig:cost-accuracy-frontier}
\end{figure}
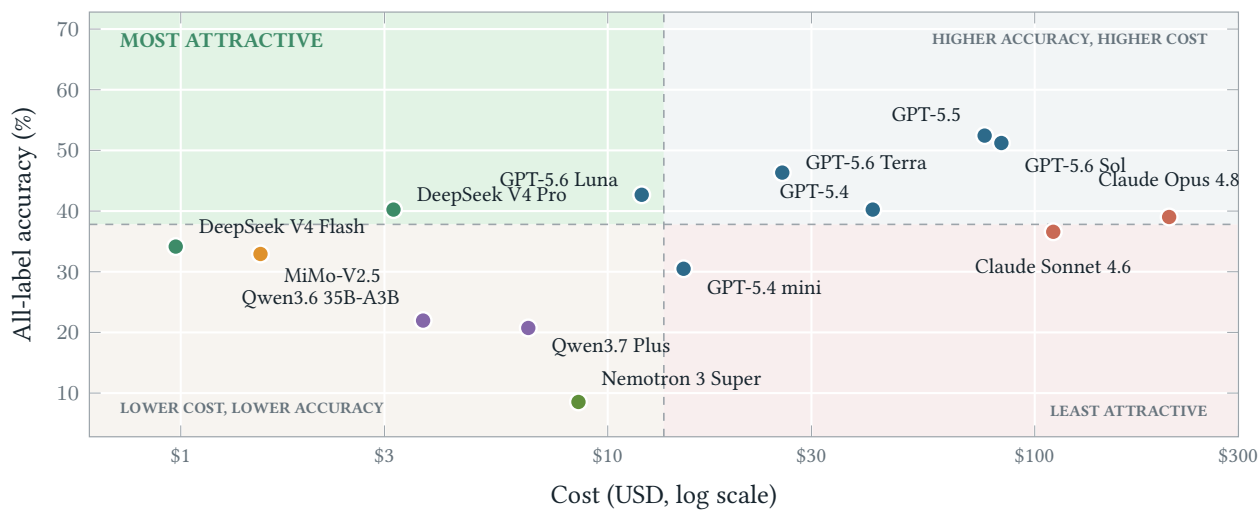
\FloatBarrier{}

\section{Introduction}\label{sec:introduction}

One of the most time-consuming and important parts of real-world buy-side equity analysis is rapidly assessing new events and news flow for potential market-moving information \citep{brown2016buyside,boudoukh2019information}. This task is challenging for a number of reasons: (1) it requires deep knowledge of what historical information has already been released \citep{tetlock2011stale}, (2) it often requires complex financial analysis to disentangle the underlying change \citep{asquith2005information,tetlock2008morethanwords}, (3) it is highly time-sensitive \citep{patell1984intraday}, (4) it can cover any aspect of a company's business or related markets, (5) news flow occurs in large volumes across every asset \citep{jeon2022news,blankespoor2020processing,hirshleifer2009distraction}, and (6) a tiny fraction of overall news flow should be flagged \citep{boudoukh2019information,jeon2022news}. Most importantly, it requires something akin to financial ``taste'' where experts can rapidly identify subtle changes that imply significant shifts in company performance or sentiment \citep{brown2016buyside,mikhail2003experience,tetlock2008morethanwords}. This task is also becoming increasingly difficult as AI-assisted analysis and reporting drive growth in the volume of new information. These features make it both a highly challenging test for AI agents and a potentially crucial use case.

In addition to the core task of identifying significant news flow, assessing the impact of new events also forms part of many other important financial tasks. For example, in order to update a model, an analyst will often need to review the latest developments to inform changes to model assumptions and projections \citep{asquith2005information,kothari2016analysts}. In order to value a stock or assess a potential acquisition, analysts will often also consider whether there are any recent relevant market, economic, or company events that should inform expectations \citep{kothari2016analysts,demirakos2004valuation}.

\begin{figure}[!htbp]
    \centering
    \includegraphics[width=0.88\textwidth]{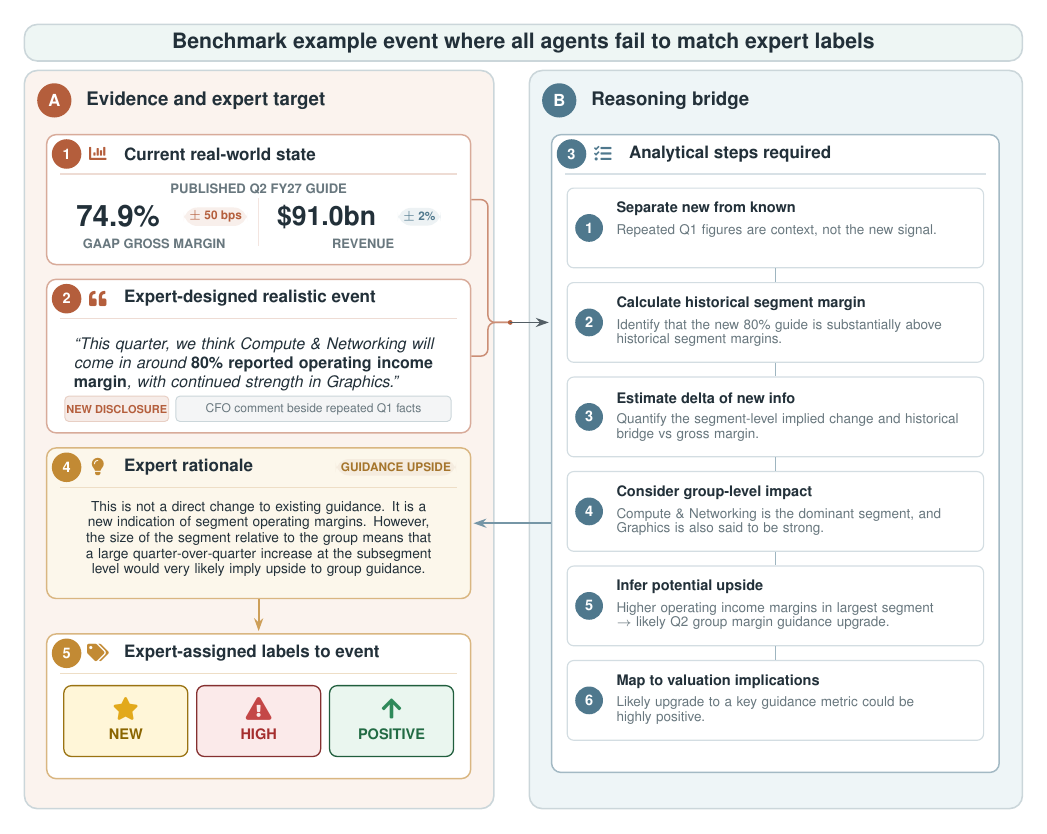}
    \caption{Example of a Frontier Financial Judgement benchmark case that requires a segment-level profitability disclosure to be translated into likely consolidated guidance upside.}\label{fig:benchmark-event-example}
\end{figure}
\FloatBarrier{}

Therefore, assessing the impact of new company news flow and events captures many of the core skills an expert equity analyst requires. However, evaluating AI agents on this task is also problematic. Valuation-relevant real-world news flow is sparse, highly clustered, and suffers from the look-ahead problem \citep{boudoukh2019information,hirshleifer2009distraction,glasserman2024lookahead}. In this work, we introduce a novel approach for robustly evaluating AI agents' capabilities on this task.

To build Frontier Financial Judgement, we collaborate with expert equity analysts to design realistic synthetic news flow grounded in current real-world company reporting. This allows us to control the difficulty, type, and novelty of the event while ensuring the event is a plausible development that could actually occur. We then mix synthetic news flow with live articles and documents to replicate the conditions an AI agent would often face in real-world usage. By combining purely synthetic events with actual articles, we allow agents to access normal web search tools while still ensuring that there is minimal risk the correct answer will leak from online sources \citep{jacovi2023contamination,white2025livebench}.

\section{Related work}\label{sec:related-work}

\subsection{Financial LLM and agent evaluations}

Historical finance LLM evaluations often used question answering to assess the financial knowledge and reasoning capabilities of LLMs. Earlier work primarily focused on static document collections; for example, FinanceBench evaluates open-book question answering over public company filings \citep{islam2023financebench}.

More recently, a number of benchmarks have evaluated LLM-based agents on broader financial research tasks with access to search engines, SEC filings and specialist tools \citep{bigeard2025financeagent,hu2025finsearchcomp,wang2026bigfinancebench}. Related work has also measured document and passage retrieval, or used structured rubrics and logic trees to assess intermediate research steps \citep{choi2025finagentbench,sun2025finresearchbench}. Most recently, DiligenceBench evaluates agents, defined by their model and harness, on open-ended equity-research questions, with answers scored against dense task-specific rubrics \citep{musaeus2026diligencebench}.

\subsection{Financial news and point-in-time evaluations}

Financial news evaluations have most commonly focused on sentiment classification, event extraction, or subsequent stock movement \citep{zhou2021tradetheevent,lefort2025finmarba}. For example, \citet{chen2024efsa} introduce an event-level financial sentiment task which jointly extracts the relevant company, event and sentiment from news articles. More recent Retrieval-Augmented Generation (RAG) work has extended this to temporally constrained financial analysis using news, filings, prices and other data \citep{zhu2025fintmmbench,zhao2026pointintime}. Point-in-time financial RAG is particularly relevant as it evaluates news-triggered event-impact prediction using only contemporaneous evidence. However, it uses subsequently realised market returns as labels, which may conflate event interpretation with wider market conditions.

The closest prior work to our label structure is by \citet{bluteau2025esgnews}, which uses LLMs to annotate ESG news for novelty, relevance, materiality and severity. Separately, ExAnte evaluates whether LLMs comply with fixed information cutoffs and measures temporal leakage \citep{liu2026exante}. These works motivate explicit novelty and materiality labels and a reproducible evidence cutoff, but do not jointly evaluate prior-information retrieval, valuation impact and directional reasoning using web-enabled agents.

\subsection{Financial news filtering}

Our work with AI agents builds on the established finance literature investigating whether human investors distinguish genuinely new information from repeated or recombined news. In particular, \citet{tetlock2011stale} defines the staleness of company news using textual similarity to prior articles and finds that investors continue to react to stale information, with subsequent return reversals. More recently, \citet{fedyk2023oldnews} show that even finance professionals can struggle to identify old information when it is recombined from multiple sources, and that these recombinations generate larger price responses than direct reprints. These studies establish that filtering previously disclosed information is an economically meaningful component of financial analysis, rather than a purely linguistic novelty task.

\subsection{Financial misinformation and verification}

Finally, this work also intersects with previous research investigating financial claim verification and misinformation detection. FinDVer evaluates evidence-grounded claim verification over long financial documents, whilst other work introduces classification and explanation benchmarks for financial misinformation \citep{zhao2024findver,liu2025fmdllama}. More recently, RFC-Bench uses minimally perturbed real financial news to test counterfactual changes, and AuditFraudBench evaluates misleading disclosure narratives using company filings and regulatory evidence \citep{jiang2026rfcbench,liu2026auditfraudbench}.

These evaluations are relevant because financial articles can remain plausible whilst being misleading through accounting, scope, causal, or temporal framing. However, determining whether an article is false is distinct from determining whether it contains new information that is material to a company's valuation. An article may be factually correct whilst recycling prior disclosures, reporting economically irrelevant details, or presenting indirect industry information with limited read-through. Therefore, there remains an important gap for an agent benchmark that jointly evaluates novelty, materiality and directional impact at a fixed evidence cutoff, including under realistic information overload.

\begin{figure}[!htbp]
    \centering
    \includegraphics[width=\textwidth]{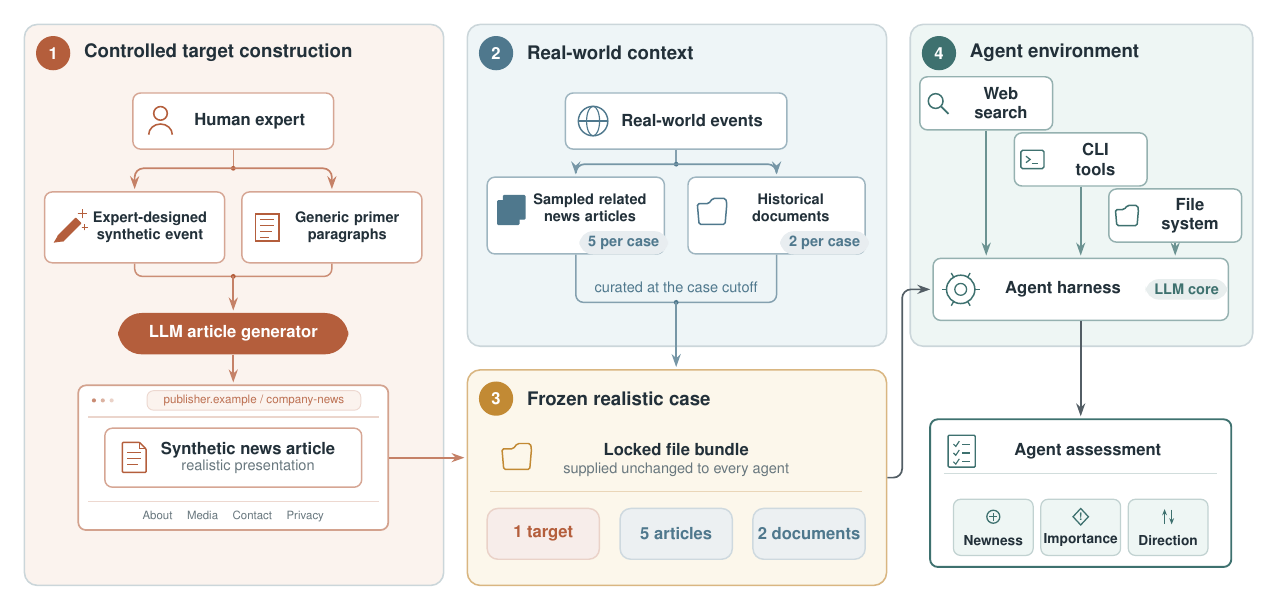}
    \caption{Overview of the benchmark setup.}\label{fig:realistic-setting-flow}
\end{figure}

\section{Methods}\label{sec:methods}

Assessing real-world financial news can require a wide spectrum of financial analysis, company context, and in-depth research. Online news articles and press releases are also often adversarial, repeating old disclosures, containing subtle wording or factual changes, overstating significance, or drawing tenuous conclusions. This makes it a hard and very general task for financial AI agents.

However, evaluating agents on this task is particularly challenging because real valuation-relevant news is sparse, very hard to isolate from other similar stories, and often ambiguous, making a live real-world benchmark intractable. To address these issues, in this work we use a combination of live news and synthetic expert-designed articles. This allows us to replicate very closely what an agent would face in the real world, but also to control the type, quantity, and difficulty of the news flow on which agents are evaluated.

\subsection{Human expert event generation}\label{sec:expert-events}

We ask professional hedge fund and equity analysts to design plausible company-specific events. For each event, the expert provides a description of the underlying business or market development, supporting evidence, gold labels, and a label rationale.

Experts assign three labels. \textit{Information new} indicates whether the article contains a factual or substantive qualitative development that was not previously public. \textit{Expected importance} is the expected impact of the information on company valuation: none, low, medium, or high. \textit{Direction} records whether the expected effect is positive, negative, neutral, or unclear. For genuine boundary cases, experts can specify more than one accepted importance or direction label. The expert-designed events are created to represent common but challenging forms of financial news; see \Cref{tab:nvidia-example-events} for examples of synthetic events relating to \CompanyA\@.

\begin{table}[!htbp]
    \centering
    \caption{Illustrative \CompanyA{} events used in Frontier Financial Judgement. The examples span recycled disclosures, roadmap changes, interpretation of opportunity size, and financially material read-through from detailed guidance.}\label{tab:nvidia-example-events}
    \begingroup
    \small
    \setlength{\tabcolsep}{4.5pt}
    \renewcommand{\arraystretch}{1.22}
    \begin{tabularx}{\textwidth}{
        >{\hsize=.80\hsize\linewidth=\hsize\raggedright\arraybackslash}X
        >{\hsize=1.25\hsize\linewidth=\hsize\raggedright\arraybackslash}X
        >{\hsize=.70\hsize\linewidth=\hsize\raggedright\arraybackslash}X
        >{\hsize=1.25\hsize\linewidth=\hsize\raggedright\arraybackslash}X
    }
        \toprule
        \rowcolor{tbink}
        \color{white}\textbf{Synthetic event} &
        \color{white}\textbf{Details} &
        \color{white}\textbf{Labels} &
        \color{white}\textbf{Label rationale} \\
        \midrule

        \rowcolor{tbrow}
        \textbf{Margin variability restated} &
        A new CFO interview says memory pricing could make second-quarter gross margin fluctuate by up to half a percentage point. &
        \textbf{New:} \textcolor{tbmuted}{No}\newline
        \textbf{Imp.:} None\newline
        \textbf{Dir.:} Neutral &
        The same gross-margin variability, up to half a percentage point, was already contained in first-quarter guidance and prior CFO commentary; fresh article framing adds no new valuation-relevant information. \\
        \midrule

        \textbf{Rubin Ultra timing slips} &
        The CEO places Rubin Ultra availability in early 2028, versus the previously communicated second-half 2027 window. &
        \textbf{New:} \textcolor{tbblue}{Yes}\newline
        \textbf{Imp.:} High\newline
        \textbf{Dir.:} \textcolor{tbred}{Negative} &
        The new date implies a possible one-to-two-quarter delay, weakening the annual product-cadence thesis that supports \CompanyA's competitive position against custom accelerators. \\
        \midrule

        \rowcolor{tbrow}
        \textbf{Vera folded into Rubin opportunity} &
        Investor relations says the standalone Vera CPU opportunity should be viewed within the broader Vera Rubin outlook, rather than stacked as a separate incremental pool. &
        \textbf{New:} \textcolor{tbblue}{Yes}\newline
        \textbf{Imp.:} Medium\textsuperscript{a}\newline
        \textbf{Dir.:} \textcolor{tbred}{Negative} &
        It was previously stated by \CompanyA{} management that the Vera CPU opportunity was standalone and in addition to Rubin estimates. The change therefore reduces overall stated opportunity size, although it could partly reflect presentation rather than final economics. \\
        \midrule

        \textbf{Subsegment margin signals guide upside} &
        The CFO says Compute \& Networking operating margin could reach approximately 80\% in the second quarter, alongside continued strength in Graphics. &
        \textbf{New:} \textcolor{tbblue}{Yes}\newline
        \textbf{Imp.:} High\newline
        \textbf{Dir.:} \textcolor{tbblue}{Positive} &
        An 80\% operating margin in \CompanyA's largest reporting group would likely imply consolidated second-quarter performance above the company's existing top-level guide. \\
        \midrule

        \rowcolor{tbrow}
        \textbf{Positive commentary recycled} &
        An article presents familiar claims about frontier-model share, Vera CPU demand, LPX, and AI-factory expansion as fresh upside. &
        \textbf{New:} \textcolor{tbmuted}{No}\newline
        \textbf{Imp.:} None\newline
        \textbf{Dir.:} Neutral &
        Each theme had already been discussed publicly. There is no new numerical guidance, customer commitment, product-schedule change, or financial update. \\
        \bottomrule
    \end{tabularx}

    \vspace{3pt}
    \begin{minipage}{\textwidth}
        \footnotesize\textit{Notes:} Labels report information newness, expected importance, and expected direction; \textsuperscript{a} high importance is also accepted for this event because the disclosure may be interpreted as a reduction in total opportunity rather than only a reporting change.
    \end{minipage}
    \endgroup
\end{table}

To construct the final articles, we combine the expert event and evidence with generic primer paragraphs. An LLM renders these inputs in different source styles and article formats, including breaking news, market commentary, and analysis notes. Finally, we apply web-page chrome templates, which add realistic surrounding content, such as navigation, reader controls, advertisements, and footers, so that the inputs closely resemble articles collected from the web.

\subsection{Benchmark setup}\label{sec:realistic-setup}

In order to closely replicate noisy real-world conditions, we place each synthetic article within a larger company-specific information bundle and use a short, realistic human instruction. Each case contains the labelled synthetic article, five recently collected real articles, and two historical company documents, all of which the agent must assess. This reflects how agents will often have to filter many pieces of news at once, without highly detailed instructions, and will often do so as part of a longer task (e.g.\ updating model assumptions).

\subsection{Dataset}\label{sec:dataset}

Our v1 benchmark dataset includes 82 synthetic events and a total of 656 items to assess. We focus initially on semiconductor supply-chain companies, including ASML, NVIDIA, Ciena, and Infineon. To avoid any risk of confusion with real events, we have removed company names from the examples in this paper; however, the real company names are provided to the agents during benchmark runs.

This dataset is designed to cover the full range of common news flow and events; representative examples are shown in \Cref{tab:news-flow-types}. For this version of the benchmark, all cases and live data were frozen as of 17 July 2026.

\begin{table}[!htbp]
    \centering
    \caption{Representative synthetic financial-news categories in Frontier Financial Judgement, with abbreviated examples from expert-designed events.}\label{tab:news-flow-types}
    \begingroup
    \small
    \setlength{\tabcolsep}{4.5pt}
    \renewcommand{\arraystretch}{1.12}
    \begin{tabularx}{\textwidth}{
        >{\hsize=.72\hsize\linewidth=\hsize\raggedright\arraybackslash}X
        >{\hsize=1.08\hsize\linewidth=\hsize\raggedright\arraybackslash}X
        >{\hsize=1.20\hsize\linewidth=\hsize\raggedright\arraybackslash}X
    }
        \toprule
        \rowcolor{tbink}
        \color{white}\textbf{Type of synthetic news} &
        \color{white}\textbf{Example} &
        \color{white}\textbf{Challenge} \\
        \midrule

        \rowcolor{tbrow}
        \textbf{Guidance restatements} &
        \CompanyB{} calls 43\% full-year gross margin consistent with its original forecast, despite its latest 44.5--45\% guide. &
        Track the guidance chronology: this is an implicit 150--200 basis-point cut despite continuity framing. \\
        \midrule

        \textbf{Economic news} &
        Taiwan electronics exports fall 6\%, while AI-relevant integrated-circuit exports rise 5\%. &
        Decompose the aggregate and map company exposure; the headline points in the wrong direction. \\
        \midrule

        \rowcolor{tbrow}
        \textbf{Restatement of previous disclosure} &
        An article presents \CompanyA's previously discussed frontier-AI share, Vera CPU, LPX and AI-factory opportunities as fresh upside. &
        Recognise old disclosures repackaged as news or combined with unsubstantiated opinion; new framing is not new information. \\
        \midrule

        \textbf{Short-seller claims} &
        A short seller argues \CompanyA's disclosed inventory build shows channel stuffing and delayed hyperscaler orders. &
        Separate public facts from unsupported causal claims and require new evidence before assigning valuation relevance. \\
        \midrule

        \rowcolor{tbrow}
        \textbf{Company updates} &
        \CompanyA{} repeats its second-quarter outlook but changes the non-GAAP margin bridge via fiscal 2025 add-backs. &
        Find the incremental detail in recycled guidance and translate an accounting bridge into earnings. \\
        \midrule

        \textbf{Product timeline shifts} &
        \CompanyB{} moves WaveLogic 7 customer sampling from 2027 to 2028. &
        Compare the new window with the prior roadmap and competing DSP timelines, while distinguishing sampling from commercial shipments. \\
        \midrule

        \rowcolor{tbrow}
        \textbf{Industry reports} &
        An industry research house reports quantities, prices and discounts for ten \CompanyA{} Vera CPU customer contracts. &
        Apply the discounts, aggregate net value to \$28.9 billion and compare it with the prior \$20 billion standalone-CPU expectation. \\
        \midrule

        \textbf{Customer contract changes} &
        A customer replaces \texteuro2.1 billion of \CompanyD{} High-NA orders with \texteuro1.25 billion of Low-NA systems and \texteuro150 million of upgrades. &
        Net removed and retained content; supplier continuity masks a \texteuro700 million reduction. \\
        \midrule

        \rowcolor{tbrow}
        \textbf{Read-across from competitors} &
        \CompetitorA{} targets volume MI450 rack shipments in the second half of 2026 with major customer commitments. &
        Translate competitor news into \CompanyA{} share and estimate implications without confusing market growth with company benefit. \\
        \midrule

        \textbf{Changes in outlook wording} &
        \CompanyC{} changes its Segment Result Margin outlook from ``around 20\%'' to ``up to 20\%''. &
        Retrieve the prior wording: one operator turns an approximate target into a ceiling amid an otherwise unchanged outlook. \\
        \bottomrule
    \end{tabularx}
    \endgroup
\end{table}

\FloatBarrier{}

\subsection{Scoring}\label{sec:scoring}

We report accuracy separately for newness, expected importance, and direction. We additionally report \textit{atomic accuracy}, the proportion of these individual classifications that are correct, and \textit{all-label accuracy}, which requires all three classifications for an event to be correct. A prediction is counted as correct when it matches either the primary expert label or an accepted secondary label for a genuine boundary case. Missing or malformed outputs remain in the benchmark denominator and are counted as incorrect.

The live articles and historical documents are curated distractors rather than individually expert-labelled negative examples. We therefore report an approximate false-positive proxy: the proportion of parseable distractor decisions that an agent classifies as both new and more important than none. We report this overall and separately for live articles and historical documents.

Accuracy alone is insufficient for assessing a practical news-filtering agent. Financial news often needs to be processed at high volume and speed for the output to remain useful. We therefore also report valid-output counts, token use, cost, and web-search volume. These secondary metrics distinguish agents that are accurate but too slow or expensive for current large-scale use and help identify how much external research is required to reach an answer.

\subsection{Agent setup and harness}\label{sec:agent-setup}

We treat the complete agent configuration, comprising the model, harness, reasoning configuration, prompt, and tool policy, as the evaluated unit. Agents can use web search to investigate company background and prior public information. One large advantage of using synthetic events is that there is minimal risk of related stories surfacing through web search and revealing the answer.

We use Harbor \citep{harborframework2026} to run each case in an isolated container and capture the answer, execution logs, and agent trajectory. For every article or document, the agent returns the three labels, a rationale, and supporting evidence in a schema-validated JSON file. We use medium reasoning across all models to balance cost and speed.

\section{Results}\label{sec:results}

We evaluate 14 agents on the same 82 realistic cases. We find that no agent can reliably replicate overall expert human judgement: the highest all-label accuracy is 52.4\%. Treating each label as a separate classification, we see higher scores but still under 75\% across all agents.

Given the importance of false positives to any real-world agent deployment, we also find that accurately classifying the synthetic news does not necessarily lead to a lower false-positive rate. For example, GPT-5.4 and Claude Opus 4.8 perform similarly on the target events, but Claude Opus has approximately five times the false-positive rate on the surrounding live articles (35\% versus 7\%).


\begin{table*}[!htbp]
    \centering
    \caption{Headline performance and operational results on the 82-case realistic benchmark. Agents are ordered by all-label accuracy within access group.}\label{tab:headline-results}
    \begingroup
    \small
    \setlength{\tabcolsep}{3.6pt}
    \renewcommand{\arraystretch}{1.16}
    \begin{tabularx}{\textwidth}{>{\raggedright\arraybackslash}Xrrrrrrr}
        \toprule
        \rowcolor{tbink}
        \color{white}\textbf{Agent} & \color{white}\textbf{All labels (\%)} & \color{white}\textbf{Atomic (\%)} & \color{white}\textbf{Valid} & \color{white}\textbf{Cost (\$)} & \color{white}\textbf{Input tok.} & \color{white}\textbf{Output tok.} & \color{white}\textbf{Searches} \\
        \midrule
        \rowcolor{tbrow}\multicolumn{8}{l}{\textbf{Closed-weight agents}} \\
        GPT-5.5 & \textbf{52.4} & \textbf{71.1} & 82/82 & 76.34 & 32.66M & 645k & 945 \\
        \rowcolor{tbrow}
        GPT-5.6 Sol & 51.2 & 69.1 & 82/82 & 83.55 & 62.36M & 644k & 1,037 \\
        GPT-5.6 Terra & 46.3 & 63.8 & 82/82 & 25.63 & 35.33M & 433k & 563 \\
        \rowcolor{tbrow}
        GPT-5.6 Luna & 42.7 & 63.4 & 82/82 & 12.01 & 40.71M & 538k & 657 \\
        GPT-5.4 & 40.2 & 59.3 & 82/82 & 41.75 & 29.83M & 771k & 2,041 \\
        \rowcolor{tbrow}
        Claude Opus 4.8 & 39.0 & 59.8 & 82/82 & 206.46 & 50.29M & 1.07M & 206 \\
        Claude Sonnet 4.6 & 36.6 & 63.0 & 82/82 & 110.52 & 54.55M & 1.47M & 433 \\
        \rowcolor{tbrow}
        GPT-5.4 mini & 30.5 & 54.1 & 81/82 & 15.06 & 32.52M & 1.21M & 2,115 \\
        \midrule
        \rowcolor{tbrow}\multicolumn{8}{l}{\textbf{Open-weight agents}} \\
        DeepSeek V4 Pro & \textbf{40.2} & 57.7 & 82/82 & 3.15 & 42.41M & 555k & 578 \\
        \rowcolor{tbrow}
        DeepSeek V4 Flash & 34.1 & \textbf{58.1} & 81/82 & 0.97 & 40.58M & 482k & 570 \\
        MiMo-V2.5 & 32.9 & 56.9 & 81/82 & 1.54 & 68.57M & 486k & 1,120 \\
        \rowcolor{tbrow}
        Qwen3.6 35B-A3B & 22.0 & 43.9 & 78/82 & 3.70 & 37.42M & 644k & 737 \\
        Qwen3.7 Plus & 20.7 & 49.2 & 77/82 & 6.52 & 44.30M & 528k & 624 \\
        \rowcolor{tbrow}
        Nemotron 3 Super & 8.5 & 15.9 & 28/82 & 8.54 & 95.11M & 428k & 594 \\
        \bottomrule
    \end{tabularx}
    \vspace{3pt}
    \begin{minipage}{\textwidth}
        \footnotesize\textit{Notes:} Atomic accuracy is the proportion of the 246 newness, importance, and direction classifications that are correct; all-label accuracy requires all three labels for a case to be correct. Malformed or missing outputs remain in the 82-case denominator. Cost, token, and search totals include the single attempt represented by each score: valid completed outputs and agent-completed invalid-format outputs scored as incorrect. Failed and superseded attempts are excluded. API-billed agents use reconciled request costs; subscription-access agents use harness-reported usage-equivalent costs. Search-service charges and local compute are excluded.
    \end{minipage}
    \endgroup
\end{table*}

\subsection{Model performance}\label{sec:model-performance}

As shown in \Cref{tab:headline-results}, GPT-5.5 is the highest-scoring agent, achieving 71.1\% atomic accuracy and 52.4\% all-label accuracy. GPT-5.6 Sol follows closely at 69.1\% and 51.2\%, respectively. Performance then falls away: the next-best agent reaches 63.8\% atomic and 46.3\% all-label accuracy. Qwen3.6 35B-A3B reaches 43.9\% atomic and 22.0\% all-label accuracy with 78 valid answers, while Nemotron 3 Super reaches 15.9\% and 8.5\% with only 28 valid answers. Nemotron's result therefore reflects substantial output and execution unreliability as well as classification errors. Even the strongest agent is fully correct on only around half of the events, despite getting more than two-thirds of the individual labels right. \Cref{fig:shared-margin-ceiling-failure} shows one example of an all-label failure shared by every agent.


\begin{figure*}[!htbp]
    \centering
    \begingroup
    \begin{tbexamplebox}{Shared failure: one operator changes the guidance}
    \setlength{\tabcolsep}{6pt}
    \renewcommand{\arraystretch}{1.24}
    \begin{tabularx}{\linewidth}{>{\centering\arraybackslash}X>{\centering\arraybackslash}X>{\centering\arraybackslash}X}
        \rowcolor{tbexamplehead}
        \textbf{Previous guidance} & \textbf{New article info} & \textbf{What changed} \\
        Segment Result Margin \textbf{around 20\%} & Segment Result Margin \textbf{up to 20\%} & An approximate target becomes a ceiling \\
        \rowcolor{tbexamplehead}
        \textbf{Prop. Correct} & \textbf{Majority response} & \textbf{Expert assessment} \\
        \textbf{0/14 (0\%)} & \textcolor{tbred}{\textbf{11/14: Not new\textsuperscript{a}, none\textsuperscript{b}, neutral\textsuperscript{c}}} & \textcolor{tbblue}{\textbf{New\textsuperscript{a}, high\textsuperscript{b}}}, \textcolor{tbred}{\textbf{negative\textsuperscript{c}}} \\
        \midrule
        \multicolumn{3}{>{\raggedright\arraybackslash}p{0.96\linewidth}}{\textbf{Common failure mode.} Eleven agents matched the repeated company, period, metric and number to the earlier disclosure, then treated the article as recycled. The remaining outputs also miss the negative direction; one identifies the change and assigns an accepted importance label but predicts positive direction.} \\
    \end{tabularx}
    \par\vspace{2pt}
    {\centering\footnotesize\textit{\textsuperscript{a} Information newness \quad \textsuperscript{b} Level of importance \quad \textsuperscript{c} Direction}\par}
    \end{tbexamplebox}
    \endgroup
    \caption{A shared all-label failure on \CompanyC's margin guidance. None of the 14 agents matches the complete expert assessment; 11 return the majority not-new, none and neutral response.}\label{fig:shared-margin-ceiling-failure}
\end{figure*}

Recorded cost is not monotonic with accuracy. DeepSeek V4 Flash comes within 1.7 percentage points of Claude Opus 4.8 on atomic accuracy (58.1\% versus 59.8\%) at recorded costs of \$0.97 and \$206.46, respectively. DeepSeek V4 Pro reaches the same 40.2\% all-label accuracy as GPT-5.4 at \$3.15 rather than \$41.75, although its atomic accuracy is 1.6 points lower. Conversely, Qwen3.7 Plus costs more than MiMo-V2.5 while scoring 7.7 points lower atomically. Qwen3.6 costs only \$3.70 but reaches 43.9\% atomic accuracy.

Research behaviour also differs substantially without following the accuracy ranking. GPT-5.4 mini issues 2,115 searches, more than twice GPT-5.5's 945, but scores 17.0 points lower atomically. At almost the same accuracy, Claude Opus 4.8 uses only 206 searches compared with GPT-5.4's 2,041. Token use shows the same pattern: Nemotron consumes the most input tokens at 95.11 million across only 59 score-bearing executions, yet produces just 28 valid answers. GPT-5.5 leads the benchmark using 32.66 million input tokens, while Claude Sonnet 4.6 produces the most output tokens at 1.47 million without reaching the leading accuracy tier. The observed totals therefore provide no evidence that longer trajectories or more extensive search alone improve event classification.

\FloatBarrier{}

\subsection{Label-level performance}\label{sec:label-performance}


\begin{table*}[!htbp]
    \centering
    \caption{Accuracy by target label and strict all-label accuracy, reported as percentages and split by the expert newness label.}\label{tab:label-performance}
    \begingroup
    \small
    \setlength{\tabcolsep}{5.2pt}
    \renewcommand{\arraystretch}{1.16}
    \begin{tabularx}{\textwidth}{>{\raggedright\arraybackslash}Xrrrrr}
        \toprule
        \rowcolor{tbink}
        \color{white}\textbf{Agent} & \color{white}\textbf{Newness} & \color{white}\textbf{Importance} & \color{white}\textbf{Direction} & \color{white}\textbf{All: new} & \color{white}\textbf{All: not new} \\
        \midrule
        \rowcolor{tbrow}\multicolumn{6}{l}{\textbf{Closed-weight agents}} \\
        GPT-5.5 & \textbf{80.5} & \textbf{62.2} & \textbf{70.7} & \textbf{50.0} & 61.1 \\
        \rowcolor{tbrow}
        GPT-5.6 Sol & 76.8 & 61.0 & 69.5 & 43.8 & 77.8 \\
        GPT-5.6 Terra & 69.5 & 58.5 & 63.4 & 35.9 & \textbf{83.3} \\
        \rowcolor{tbrow}
        GPT-5.6 Luna & 75.6 & 58.5 & 56.1 & 34.4 & 72.2 \\
        Claude Sonnet 4.6 & 76.8 & 50.0 & 62.2 & 32.8 & 50.0 \\
        \rowcolor{tbrow}
        Claude Opus 4.8 & 68.3 & 51.2 & 59.8 & 29.7 & 72.2 \\
        GPT-5.4 & 70.7 & 51.2 & 56.1 & 28.1 & \textbf{83.3} \\
        \rowcolor{tbrow}
        GPT-5.4 mini & 62.2 & 47.6 & 52.4 & 21.9 & 61.1 \\
        \midrule
        \rowcolor{tbrow}\multicolumn{6}{l}{\textbf{Open-weight agents}} \\
        DeepSeek V4 Flash & 65.9 & 51.2 & \textbf{57.3} & 28.1 & \textbf{55.6} \\
        \rowcolor{tbrow}
        DeepSeek V4 Pro & \textbf{68.3} & \textbf{53.7} & 51.2 & \textbf{39.1} & 44.4 \\
        MiMo-V2.5 & \textbf{68.3} & 50.0 & 52.4 & 35.9 & 22.2 \\
        \rowcolor{tbrow}
        Qwen3.7 Plus & 64.6 & 36.6 & 46.3 & 15.6 & 38.9 \\
        Qwen3.6 35B-A3B & 47.6 & 40.2 & 43.9 & 12.5 & \textbf{55.6} \\
        \rowcolor{tbrow}
        Nemotron 3 Super & 18.3 & 14.6 & 14.6 & 1.6 & 33.3 \\
        \bottomrule
    \end{tabularx}
    \vspace{3pt}
    \begin{minipage}{\textwidth}
        \footnotesize\textit{Notes:} Values are percentages. Each individual label uses all 82 cases; invalid outputs score zero. Predictions matching an expert-accepted secondary importance or direction label are correct. The strict split uses 64 gold-new and 18 gold-not-new events.
    \end{minipage}
    \endgroup
\end{table*}

Newness is the strongest individual label across the evaluated agents, with 749/1,148 decisions correct (65.2\%), compared with 620/1,148 for direction (54.0\%) and 563/1,148 for expected importance (49.0\%). Every agent has lower importance accuracy than newness accuracy (\Cref{tab:label-performance}). The central difficulty is therefore not only recognising that an article contains a new fact, but calibrating how much that fact should change valuation-relevant beliefs.

This difficulty is more apparent under the joint measure. Across all agent-event decisions, all-label accuracy is 262/896 (29.2\%) on the 64 genuinely new events, compared with 146/252 (57.9\%) on the 18 not-new events. Most not-new events resolve to the simpler joint output of not new, no importance and neutral direction. A new event additionally requires the agent to translate the factual change into an importance and direction judgement. GPT-5.5 has the highest all-label accuracy on new events at 50.0\%, while no other agent exceeds 43.8\%. \Cref{fig:run-rate-direction-trap} illustrates how failure to get one of the expert labels often highlights a fundamental misunderstanding of the important change.


\begin{figure*}[!htbp]
    \centering
    \begingroup
    \begin{tbexamplebox}{Direction trap: the headline uses the wrong comparison}
    \setlength{\tabcolsep}{6pt}
    \renewcommand{\arraystretch}{1.24}
    \begin{tabularx}{\linewidth}{>{\centering\arraybackslash}X>{\centering\arraybackslash}X>{\centering\arraybackslash}X}
        \rowcolor{tbexamplehead}
        \textbf{Positive headline} & \textbf{Required comparison} & \textbf{Hidden reversal} \\
        Q1 FY27 revenue expected to rise 10\% year on year & Q1 FY27 $\approx$ \texteuro4.028bn; implied Q4 FY26 $\geq$ \texteuro4.426bn & $\geq$9\% sequential decline, versus prior guidance that Q1 should be better than Q4 \\
        \rowcolor{tbexamplehead}
        \textbf{Agent newness} & \textbf{Agent direction} & \textbf{Expert assessment} \\
        \textcolor{tbblue}{\textbf{14/14 identify it as new\textsuperscript{a}}} & \textcolor{tbred}{\textbf{14/14 predict positive\textsuperscript{c} direction}} & \textcolor{tbblue}{\textbf{New\textsuperscript{a}, high\textsuperscript{b}}}, \textcolor{tbred}{\textbf{negative\textsuperscript{c}}} \\
        \midrule
        \multicolumn{3}{>{\raggedright\arraybackslash}p{0.96\linewidth}}{\textbf{Common failure mode.} The agents followed the article's positive year-on-year framing. They did not change the comparison base to the preceding quarter, where the figures imply a decline of at least 9\% and reverse prior guidance.} \\
    \end{tabularx}
    \par\vspace{2pt}
    {\centering\footnotesize\textit{\textsuperscript{a} Information newness \quad \textsuperscript{b} Level of importance \quad \textsuperscript{c} Direction}\par}
    \end{tbexamplebox}
    \endgroup
    \caption{The \CompanyC{} fiscal Q1 2027 case separates novelty detection from directional reasoning. Every agent recognises the new forecast, but the positive year-on-year framing masks a negative sequential revision relative to the prior outlook.}\label{fig:run-rate-direction-trap}
\end{figure*}

\FloatBarrier{}

\subsection{False-positive trade-off}\label{sec:false-positive-trade-off}


\begin{table*}[!htbp]
    \centering
    \caption{Approximate false-positive rates on curated noise items, overall and by item type.}\label{tab:false-positive-rates}
    \begingroup
    \small
    \setlength{\tabcolsep}{7pt}
    \renewcommand{\arraystretch}{1.16}
    \begin{tabularx}{\textwidth}{>{\raggedright\arraybackslash}Xrrr}
        \toprule
        \rowcolor{tbink}
        \color{white}\textbf{Agent} & \color{white}\textbf{Overall} & \color{white}\textbf{Live articles} & \color{white}\textbf{Historical documents} \\
        \midrule
        \rowcolor{tbrow}\multicolumn{4}{l}{\textbf{Closed-weight agents}} \\
        GPT-5.5 & 37/574 (6.4\%) & 37/410 (9.0\%) & 0/164 (0.0\%) \\
        \rowcolor{tbrow}
        GPT-5.6 Sol & \textbf{6/574 (1.0\%)} & 5/410 (1.2\%) & 1/164 (0.6\%) \\
        GPT-5.6 Terra & 29/574 (5.1\%) & 29/410 (7.1\%) & 0/164 (0.0\%) \\
        \rowcolor{tbrow}
        GPT-5.6 Luna & 84/574 (14.6\%) & 84/410 (20.5\%) & 0/164 (0.0\%) \\
        Claude Sonnet 4.6 & 185/574 (32.2\%) & 184/410 (44.9\%) & 1/164 (0.6\%) \\
        \rowcolor{tbrow}
        Claude Opus 4.8 & 143/574 (24.9\%) & 143/410 (34.9\%) & 0/164 (0.0\%) \\
        GPT-5.4 & 33/574 (5.7\%) & 29/410 (7.1\%) & 4/164 (2.4\%) \\
        \rowcolor{tbrow}
        GPT-5.4 mini & 55/567 (9.7\%) & 55/405 (13.6\%) & 0/162 (0.0\%) \\
        \midrule
        \rowcolor{tbrow}\multicolumn{4}{l}{\textbf{Open-weight agents}} \\
        DeepSeek V4 Flash & 108/567 (19.0\%) & 108/405 (26.7\%) & 0/162 (0.0\%) \\
        \rowcolor{tbrow}
        DeepSeek V4 Pro & 99/574 (17.2\%) & 99/410 (24.1\%) & 0/164 (0.0\%) \\
        MiMo-V2.5 & 117/567 (20.6\%) & 113/405 (27.9\%) & 4/162 (2.5\%) \\
        \rowcolor{tbrow}
        Qwen3.7 Plus & 134/539 (24.9\%) & 134/385 (34.8\%) & 0/154 (0.0\%) \\
        Qwen3.6 35B-A3B & \textbf{84/546 (15.4\%)} & 84/390 (21.5\%) & 0/156 (0.0\%) \\
        \rowcolor{tbrow}
        Nemotron 3 Super & 31/196 (15.8\%) & 28/140 (20.0\%) & 3/56 (5.4\%) \\
        \bottomrule
    \end{tabularx}
    \vspace{3pt}
    \begin{minipage}{\textwidth}
        \footnotesize\textit{Notes:} A noise decision is counted when an agent labels the item as both new and more important than none. This is an approximate escalation proxy rather than a conventional gold-labelled false-positive rate. Denominators include only parseable decisions from valid answers; repeated pool items are counted each time they occur in a case.
    \end{minipage}
    \endgroup
\end{table*}

Approximate false-positive behaviour varies substantially and is not determined by target-event accuracy (\Cref{tab:false-positive-rates}). GPT-5.6 Sol combines near-leading target performance with the lowest overall rate, at 1.0\%, compared with 6.4\% for GPT-5.5. GPT-5.6 Terra and Luna have almost identical atomic accuracy (63.8\% and 63.4\%) but false-positive rates of 5.1\% and 14.6\%. Similarly, Claude Opus 4.8 and GPT-5.4 differ by only one correct atomic classification, but their false-positive rates are 24.9\% and 5.7\%.

The false positives are concentrated almost entirely in live articles. Across all agents, 1,132/5,410 article decisions (20.9\%) are escalated as both new and more important than none, compared with 13/2,164 historical-document decisions (0.6\%). Thus, 1,132 of the 1,145 estimated false positives arise from articles. Agents can usually recognise that an old filing or company report is not current news; the harder problem here, and in the real world, is distinguishing genuinely incremental information from recently published derivative coverage, repeated announcements, and immaterial read-across.

\FloatBarrier{}

\section{Discussion}\label{sec:discussion}

Our results suggest that current frontier agents still cannot assess equity news flow to the standard of professional equity analysts. The strongest agent, GPT-5.5, matches the expert assessment on all three labels in only 52.4\% of cases. GPT-5.6 Sol reaches 51.2\% all-label accuracy and Claude Opus 4.8 reaches 39.0\%. Their higher atomic scores show that they often identify part of the correct answer, but this is insufficient for a task where novelty, importance and direction jointly determine whether and how an event should be escalated. Even the strongest agents therefore fail to reproduce the complete expert judgement on almost half of the events.

The shared failures indicate that the remaining gap is not simply one of retrieving more information. Agents overlook subtle changes within otherwise repeated disclosures, as in the shift from a margin target of \textit{around 20\%} to a ceiling of \textit{up to 20\%} in \Cref{fig:shared-margin-ceiling-failure}. They also accept favourable article framing without selecting the comparison that is financially relevant. In \Cref{fig:run-rate-direction-trap}, every agent recognises that the forecast is new, but every agent follows its positive year-on-year headline rather than calculating the negative sequential revision implied by the figures and prior guidance. Together with the consistently weaker importance results, these cases suggest that agents often fail to integrate small factual changes, historical expectations and their implied financial impact. These are failures of contextual financial judgement rather than simple document retrieval or sentiment classification.

This limitation matters both for direct news monitoring and for broader financial workflows. In a news-filtering application, missing a small but consequential change creates a false negative, while an incorrect importance or direction judgement can cause analysts to prioritise the wrong event. The same assessment is also an upstream input to updating financial models, revising valuations, assessing acquisitions and conducting wider equity research. Frontier Financial Judgement does not evaluate these longer workflows end to end, so our results do not establish their overall performance. They do, however, identify a consequential subtask on which the quality of downstream analysis may depend.

For real-world applications, accuracy is not the only factor. Latency and cost are often important barriers to practical deployment at scale. Although recorded benchmark cost and token use are imperfect proxies for actual wall-clock time, they reflect the fact that substantial agent trajectories must be completed within a realistic trading window after an article is published. Practical deployment must therefore trade off accuracy against both cost and timeliness, with \Cref{fig:cost-accuracy-frontier} showing a clear cost-accuracy frontier. Therefore, even if frontier models approach human expert judgement in the future, there will likely still be a substantial need for research into harness and model optimisation in order to meet real-world latency and cost requirements.

Finally, target accuracy and false-positive behaviour form separate components of real-world usability. GPT-5.6 Sol combines near-leading target accuracy with a 1.0\% approximate false-positive rate, whereas Claude Opus 4.8 reaches 24.9\%. The comparison between Claude Opus and GPT-5.4 is particularly instructive: they differ by only one correct atomic classification, yet GPT-5.4's approximate false-positive rate is 5.7\%. In a low-base-rate news environment, these differences can translate into substantially different volumes of irrelevant material being escalated for review. GPT-5.6 Sol provides the strongest combined result in this experiment, but no single accuracy measure captures the operational trade-off. Practical news agents should therefore be evaluated jointly on judgement accuracy, restraint, output reliability and cost. As the noise items are curated distractors rather than an individually expert-labelled negative set, these rates remain controlled estimates of noise escalation rather than population-wide false-positive rates for financial news.

\section{Limitations}\label{sec:limitations}

The primary limitation of the synthetic-event approach is that it does not reproduce the noise created by competing interpretations of the same information across different articles. Real-world news-flow filtering may therefore be even more challenging than the benchmark suggests. This first version of Frontier Financial Judgement focuses on semiconductor supply-chain companies, and its results may not generalise to other sectors. Finally, some expert judgements are inherently subjective, and other financial professionals may reasonably have assigned different labels.

\section{Conclusion}\label{sec:conclusion}

This paper introduces Frontier Financial Judgement, a benchmark for evaluating whether research agents can reproduce professional equity analyst assessments of news newness, importance and direction. Across 14 agents, the strongest result reaches only 52.4\% all-label accuracy. Even frontier agents frequently miss subtle changes, accept misleading framing or fail to integrate an event's implied financial impact. Their performance also varies substantially in cost, output reliability and restraint when presented with irrelevant material.

These findings suggest that current agents can support equity-news monitoring, but cannot yet replace professional judgement in consequential financial workflows. Practical deployment requires evaluation of complete decisions rather than isolated labels, alongside false-positive behaviour, reliability and cost. Frontier Financial Judgement provides a reproducible foundation for measuring progress on these dimensions, while future work should extend its coverage across sectors and more closely reproduce the competing interpretations found in real-world news flow.

\bibliographystyle{plainnat}
\bibliography{references}

\clearpage
\appendix
\section{Appendix}\label{app:benchmark-documentation}

\subsection{Prompts}\label{app:prompts}

\Cref{fig:realistic-prompt} reproduces the complete instruction used for the realistic benchmark. The agent receives this instruction alongside a task manifest containing the company, ticker, assessment cutoff and paths to the eight supplied items.

\begin{figure}[!ht]
    \centering
    \begin{tcolorbox}[
        enhanced,
        width=\textwidth,
        colback=tbexampleback,
        colframe=tbexampleframe,
        colbacktitle=tbexampletitle,
        coltitle=white,
        fonttitle=\sffamily\bfseries,
        title={Realistic news-flow assessment prompt},
        boxrule=0.65pt,
        arc=2.4mm,
        outer arc=2.4mm,
        boxsep=0.7mm,
        left=1.8mm,
        right=1.8mm,
        top=1.4mm,
        bottom=1.4mm,
        toptitle=1.3mm,
        bottomtitle=1.3mm,
        titlerule=0pt
    ]
    \begin{minipage}[t]{0.485\textwidth}
        \raggedright\scriptsize
        I am an analyst and have been given the following articles and documents to review so I can update the equity sales team on any new developments that could be valuation relevant for the company. They are obviously busy and know the stock well so you should only flag genuinely new valuation relevant information that could actually influence company valuation.

        \medskip
        The company and ticker to assess the articles / documents are given in \texttt{/app/task.json} as \texttt{task.company} and \texttt{task.ticker}.

        \medskip
        Please provide me with three classifications for each article / document so I can only mention / pass onto the team genuinely new valuation relevant info.

        \medskip
        \textbf{New information classification}
        \begin{itemize}[leftmargin=1.2em,itemsep=0.5pt,topsep=2pt,parsep=0pt]
            \item \texttt{information\_new = false} when the important content is equivalent to information already public before \texttt{as\_of\_utc}, independently of the supplied item itself, or the item contains no substantive factual development.
            \item \texttt{information\_new = true} only when the item contains a factual or substantive qualitative development that was not otherwise public before \texttt{as\_of\_utc}.
        \end{itemize}

        \textbf{Expected importance classification}
        \begin{itemize}[leftmargin=1.2em,itemsep=0.5pt,topsep=2pt,parsep=0pt]
            \item \texttt{none}: no novel factual delta, or the novel information is not expected to change valuation-relevant beliefs substantively.
            \item \texttt{low}: new information with a plausible but limited or uncertain connection to valuation-relevant expectations.
            \item \texttt{medium}: new information likely to matter to valuation-relevant expectations, but not clearly major on its own.
            \item \texttt{high}: new information highly likely to change valuation-relevant expectations meaningfully.
        \end{itemize}

        \textbf{Direction classification}
        \begin{itemize}[leftmargin=1.2em,itemsep=0.5pt,topsep=2pt,parsep=0pt]
            \item \texttt{positive}: the novel valuation-relevant information is likely to support or raise expectations around valuation drivers.
            \item \texttt{negative}: it is likely to weigh on or lower those expectations.
            \item \texttt{neutral}: there is no meaningful novel valuation-relevant delta, or the new information has no expected valuation impact.
            \item \texttt{unclear}: there is novel potentially valuation-relevant information, but direction cannot be determined or has materially offsetting implications.
        \end{itemize}
    \end{minipage}\hfill
    \begin{minipage}[t]{0.485\textwidth}
        \raggedright\scriptsize
        \textbf{Notes}

        \smallskip
        Read \texttt{/app/task.json} first and review \texttt{task.items[*].path}.

        \medskip
        Assess every item as of \texttt{as\_of\_utc}. Do not use information or stock price movements after \texttt{as\_of\_utc}.

        \medskip
        Some supplied material may come from third-party or internal feeds and may not appear verbatim on the open web. Use web search for background and prior public information, but do not treat the absence of an exact online match as evidence that an item is false, already known, or unimportant.

        \medskip
        Use the supplied items and allowed search tools to ground each assessment. Cite key evidence in the evidence field.

        \medskip
        Evidence from a supplied article or document uses:
        \begin{itemize}[leftmargin=1.2em,itemsep=0.5pt,topsep=2pt,parsep=0pt]
            \item \texttt{type = feed\_item}
            \item \texttt{itemId} = the relevant supplied \texttt{itemId}
        \end{itemize}

        Evidence from search results uses:
        \begin{itemize}[leftmargin=1.2em,itemsep=0.5pt,topsep=2pt,parsep=0pt]
            \item \texttt{type = web\_search\_result}
            \item include \texttt{title}, \texttt{url}, and \texttt{published\_at\_utc}; do not put native-tool internal citation handles such as \texttt{turn0search4} in \texttt{result\_id}
        \end{itemize}

        Before writing the final answer, verify that \texttt{answer.assessments[*].itemId} exactly matches the \texttt{itemIds} in \texttt{/app/task.json}, with no missing, duplicate, or extra \texttt{itemIds}.

        \medskip
        Write exactly one JSON object matching \texttt{/app/answer.schema.json} to \texttt{/logs/artifacts/answer.json}.
    \end{minipage}
    \end{tcolorbox}
    \caption{The complete realistic-setting instruction presented to each agent. Markdown headings and lists are reformatted typographically, but the wording is unchanged.}\label{fig:realistic-prompt}
\end{figure}

\FloatBarrier{}

\subsection{Example primer paragraphs}\label{app:primer-paragraphs}

Synthetic articles combine event-specific evidence with generic company background. The background paragraphs are selected from a reviewed company-specific primer library and do not determine the expert labels. \Cref{fig:nvidia-primer-examples} shows three examples from the NVIDIA library.

\begin{figure}[htbp]
    \centering
    \begin{tcolorbox}[
        enhanced,
        width=\textwidth,
        colback=white,
        colframe=tbexampleframe,
        boxrule=0.65pt,
        arc=2.4mm,
        outer arc=2.4mm,
        boxsep=0.8mm,
        left=1.8mm,
        right=1.8mm,
        top=1.5mm,
        bottom=1.5mm
    ]
        \begin{tcolorbox}[
            colback=tbexampleback,
            colframe=tbexamplehead,
            colbacktitle=tbexamplehead,
            coltitle=tbink,
            fonttitle=\sffamily\bfseries,
            title={NVIDIA's shift from graphics chips to accelerated computing platforms},
            boxrule=0pt,
            arc=0.8mm,
            left=1.4mm,right=1.4mm,top=1mm,bottom=1mm
        ]
        \small NVIDIA began as a graphics processor company, but investors now tend to analyze it as a platform supplier for accelerated computing. Its business spans data-center accelerators, networking, systems, software libraries, gaming GPUs, professional visualization, automotive computing, and robotics. The company's central position in AI infrastructure comes from combining chips with interconnects, systems design, developer software, and an ecosystem of hardware and cloud partners. That platform breadth means a single product or demand update can affect expectations for multiple parts of the business, not just one chip SKU\@.
        \end{tcolorbox}

        \begin{tcolorbox}[
            colback=tbrow,
            colframe=tbexamplehead,
            colbacktitle=tbexamplehead,
            coltitle=tbink,
            fonttitle=\sffamily\bfseries,
            title={AI factories and data-center infrastructure constraints},
            boxrule=0pt,
            arc=0.8mm,
            left=1.4mm,right=1.4mm,top=1mm,bottom=1mm,
            before skip=1.5mm
        ]
        \small Large AI clusters require more than accelerator supply. Operators need power availability, cooling capacity, high-bandwidth networking, physical space, systems integration, and software to schedule workloads across many chips. NVIDIA describes this buildout as AI factory infrastructure because the facilities convert electricity and data into model training and inference output. For investors, that makes deployment pace sensitive to construction schedules, utility connections, rack readiness, and the ability of customers to absorb complete systems.
        \end{tcolorbox}

        \begin{tcolorbox}[
            colback=tbexampleback,
            colframe=tbexamplehead,
            colbacktitle=tbexamplehead,
            coltitle=tbink,
            fonttitle=\sffamily\bfseries,
            title={Gross-margin sensitivity to product mix and supply chain},
            boxrule=0pt,
            arc=0.8mm,
            left=1.4mm,right=1.4mm,top=1mm,bottom=1mm,
            before skip=1.5mm
        ]
        \small NVIDIA's margins can be affected by product mix, system content, memory costs, packaging capacity, and the balance between chips, full systems, and software. Data-center products have often supported strong profitability, but more complex rack-scale systems can also bring different cost structures. Investors therefore pay close attention to whether new information points to changes in pricing power, supply costs, or the mix of products being shipped.
        \end{tcolorbox}
    \end{tcolorbox}
    \caption{Example generic background primers used when constructing NVIDIA synthetic articles.}\label{fig:nvidia-primer-examples}
\end{figure}

\FloatBarrier{}

\subsection{Agent setup and harness}\label{app:agent-setup}

All agents run within Harbor using the same frozen cases, answer schema, artifact capture and verification procedure. The evaluated model-specific configurations are:

\begin{itemize}[leftmargin=1.4em,itemsep=3pt,topsep=3pt]
    \item \textbf{GPT-5.4:} Harbor with the single-agent Codex adapter and Codex CLI 0.142.3; native Codex web search; OpenAI through Codex subscription access.
    \item \textbf{GPT-5.4 mini:} Harbor with the single-agent Codex adapter and Codex CLI 0.142.3; native Codex web search; OpenAI through Codex subscription access.
    \item \textbf{GPT-5.5:} Harbor with the single-agent Codex adapter and Codex CLI 0.142.3; native Codex web search; OpenAI through Codex subscription access.
    \item \textbf{GPT-5.6 Sol:} Harbor with the single-agent Codex adapter and Codex CLI 0.144.1; native Codex web search; OpenAI through Codex subscription access.
    \item \textbf{GPT-5.6 Terra:} Harbor with the single-agent Codex adapter and Codex CLI 0.144.1; native Codex web search; OpenAI through Codex subscription access.
    \item \textbf{GPT-5.6 Luna:} Harbor with the single-agent Codex adapter and Codex CLI 0.144.1; native Codex web search; OpenAI through Codex subscription access.
    \item \textbf{Claude Opus 4.8:} Harbor with the single-agent Claude Code adapter and Claude Code 2.1.206; native Claude web search; Anthropic through Claude subscription access.
    \item \textbf{Claude Sonnet 4.6:} Harbor with the single-agent Claude Code adapter and Claude Code 2.1.206; native Claude web search; Anthropic through Claude subscription access.
    \item \textbf{DeepSeek V4 Flash:} Harbor with the single-agent OpenCode adapter and OpenCode 1.18.2; the shared Exa web-search tool; OpenRouter routing pinned to DeepSeek with provider fallback disabled.
    \item \textbf{DeepSeek V4 Pro:} Harbor with the single-agent OpenCode adapter and OpenCode 1.18.2; the shared Exa web-search tool; OpenRouter routing pinned to DeepSeek with provider fallback disabled.
    \item \textbf{MiMo-V2.5:} Harbor with the single-agent OpenCode adapter and OpenCode 1.18.2; the shared Exa web-search tool; OpenRouter routing pinned to Xiaomi's FP8 endpoint with provider fallback disabled.
    \item \textbf{Qwen3.7 Plus:} Harbor with the single-agent OpenCode adapter and OpenCode 1.18.2; the shared Exa web-search tool; OpenRouter routing pinned to Alibaba with provider fallback disabled.
    \item \textbf{Qwen3.6 35B-A3B:} Harbor with the authenticated-Exa OpenCode adapter and OpenCode 1.18.2; the shared Exa web-search tool; OpenRouter routing pinned to Parasail's FP8 endpoint with provider fallback disabled.
    \item \textbf{Nemotron 3 Super:} Harbor with the authenticated-Exa OpenCode adapter and OpenCode 1.18.2; the shared Exa web-search tool; OpenRouter routing pinned to DeepInfra's BF16 endpoint with provider fallback disabled.
\end{itemize}

\subsection{Live news articles}\label{app:live-news}

The live article pool is collected immediately before benchmark cases are frozen. For each company, the Exa Search API \citep{exa2026search} receives the configured query \texttt{Latest news on \{company\} \{ticker\}}. The request uses Exa's automatic search mode to retrieve up to 50 results published during the preceding three days, including the full extracted article text.

Candidates are normalised, deduplicated and ordered newest first. Each is then reviewed for company relevance. Direct company news and customer, supplier, competitor, regulatory, sector or market news with a credible company read-through are retained. Incidental mentions, generic ticker lists, articles too thin to classify, malformed pages and stories about unrelated entities without a plausible read-through are rejected. Review stops once the candidate batch has yielded 25 relevant articles or all candidates have been reviewed. The final 25-article pool is fixed before case construction, after which five articles are selected deterministically for each realistic benchmark case.

\end{document}